# Abstraction in Belief Networks: The Role of Intermediate States in Diagnostic Reasoning

**Gregory Provan***
Institute for Decision Systems Research
4984 El Camino Real, Suite 110
Los Altos, CA 94022

## Abstract

Bayesian belief networks are being increasingly used as a knowledge representation for diagnostic reasoning. One simple method for conducting diagnostic reasoning is to represent system faults and observations only. In this paper, we investigate how having intermediate nodes—nodes other than fault and observation nodes—affects the diagnostic performance of a Bayesian belief network. We conducted a series of experiments on a set of real belief networks for medical diagnosis in liver and bile disease. We compared the effects on diagnostic performance of a two-level network consisting just of disease and finding nodes with that of a network which models intermediate pathophysiological disease states as well. We provide some theoretical evidence for differences observed between the abstracted two-level network and the full network.

## 1 THE TRADEOFF BETWEEN ACCURACY AND COST

In recent years, there has been substantial growth in interest in Bayesian belief networks (BNs) as a knowledge representation [Pearl, 1988]. There has been work in the development of effective knowledge engineering techniques, efficient inference algorithms, and increasing numbers of real-world applications of BNs [Henrion et al., 1991].

The primary goal of the work described here is to investigate how the precision of representation of BNs affects the quality of diagnosis based on the network. We view this research as a contribution towards developing an empirical and theoretical basis for knowledge engineering guidelines. These guidelines would help knowledge engineers choose the level and complexity of representation that provides the most appropriate tradeoff between accuracy and the cost of computational resources for inference, storage and of knowledge engineering.

---
*This work was supported by NSF grants #IRI91-20330 and #IRI92-10030.

The problem domain that we use in this study is medical diagnosis for hepatobiliary disorders (liver and bile diseases), using a network called CPCS-BN. In the CPCS-BN, the variables can be divided into three classes: diseases, Intermediate Pathophysiological States (IPSs) and findings. In addition, these variables are causally ordered, with diseases "causing" IPSs and IPSs "causing" findings. This causal ordering is reflected by a directed, hierarchical ordering in CPCS-BN: disease nodes precede IPS nodes, which precede finding nodes.

In this paper we abstract CPCS-BN into a two-level network containing only disease and finding nodes, and experimentally compare the performance of the three-level and the two-level networks to determine the effect on diagnostic accuracy of our mapping. The two-level simplification is structurally isomorphic to the QMR-DT network [Shwe et al., 1991; Middleton et al., 1991], and our experiments aim to test whether having more than two levels in a network—representing intermediate disease states explicitly—leads to better diagnostic performance than just having two levels. Figure 1 shows a simple case of reducing a three-level network into a two-level version of that network that does not contain IPS nodes. This two-level mapping may be viewed as extreme, but it does provide a point of reference between the QMR-DT representation and the three-level CPCS representation.

There are some important tradeoffs involved in modeling intermediate nodes, involving (a) the number and cost of probability assessments, and (b) the independence assumptions that are necessary. For the first case, the cost of knowledge engineering is roughly proportional to the number of probabilities which need to be assessed, multiplied by the average cost of each assessment. On the one hand introducing intermediate nodes can reduce the total number of probabilities which need to be specified, as we shall show in the paper. However, on the other hand, the average cost of each assesment is higher: greater knowledge engineering effort is required to specify probabilities which are typically unobservable, such as probabilities of diseases causing IPSs.[1] Given the (effective) two-level QMR KB, an additional two man-years of knowledge engineering was spent in constructing the CPCS KB [Parker and Miller, 1987], an indication of the difficulty of modeling intermediate nodes for such domains. In the construction of CPCS-BN, disease priors were obtained from data, and other probabilities were elicited from knowledge engineers; of this latter set of probabilities, disease-finding relationships (as modeled in the two-level network) are better understood than disease-IPS or IPS-finding relationships of the three-level network, and hence are simpler to assess. This is because practical experience and data on which to base probability assessments could be obtained for disease-finding tuples more easily than for disease-IPS or IPS-finding tuples.

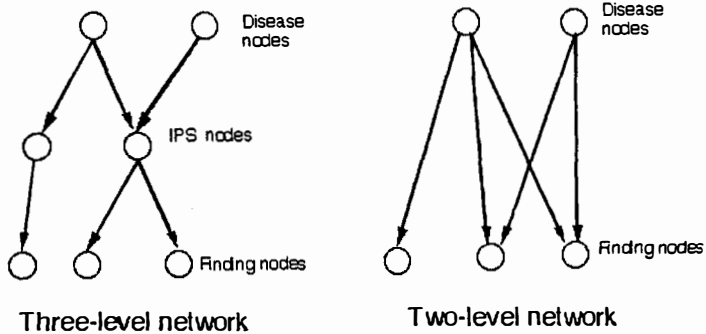

Figure 1: Network reduction of a simple network

For the second case, modeling intermediate nodes, i.e., IPSs, introduces explicit independence models into CPCS-BN, thereby reducing computational expense. Taken together, it is clear that evaluating the tradeoffs fully is a multi-attribute task: the total knowledge engineering costs are a function of the number of probabilities to be assessed, the difficulty of assessing certain types of probabilities, and the use of appropriate independence assumptions. Our study aims to explore the relationships between such knowledge engineering tradeoffs and diagnostic accuracy, and hence shed light on where abstractions in BNs are appropriate.

## 2    EXPERIMENTAL DOMAIN

In this section we briefly describe the BN we have selected for our experimental analysis, CPCS-BN. CPCS-BN supports medical diagnosis for liver and bile (hepatobiliary) diseases. We derived the network from a rich knowledge base, the CPCS system, developed by R. Parker and R. Miller [Parker and Miller, 1987] in the mid-1980s as an experimental extension of the Internist-1 knowledge base [Miller *et al.*, 1982].

---

[1]IPSs are hidden or conceptual classes for which there typically is no *direct* evidence, as there is for findings.

The CPCS KB was developed as an experimental extension of the Internist-1 KB to support patient simulation and computer aided instruction. The developers felt that these tasks required a KB with a much richer representation than that of Internist-1. CPCS is restricted to the hepatobiliary medical domain because the developers regarded the knowledge engineering task too great to convert all of Internist-1 to a richer representation based on their experience with the CPCS system.

Internist-1, and more recently QMR, contain only diseases and findings–a two-level representation.    The CPCS KB has a multilevel representation that includes diseases and findings as well as predisposing factors to diseases (PFDs) that influence disease prevalence rates, and intermediate pathophysiological states (IPSs) that mediate between diseases and findings. For example, consider a disease *acute gastritis* which involves blood loss and two findings suggestive of blood loss: *pale skin* and *low red blood cell count*. Internist-1 has direct links between this disease and the two findings. In contrast, the CPCS KB includes *anemia* (an IPS) between the disease and the two findings.

The full CPCS-BN has 448 nodes and over 900 arcs. Because inference in the complete CPCS-BN is extremely time consuming, for our experiments we used subsets of the full network comprising 42 nodes (2 diseases), 146 nodes (3 diseases) and 245 nodes (4 diseases). These subsets are described in more detail in Section 5.

## 3    BAYESIAN NETWORK REPRESENTATION

This section describes the theoretical underpinnings for our representation of causal influences. In CPCS-BN, all causal influence are leaky noisy OR, and so we describe the noisy OR representation for causal influences, followed by the leak representation for causes that are not explicitly modeled. Noisy OR and leaky-noisy OR are simplified representations for probabilistic influence that require far fewer parameters than the full conditional-probability matrix. In fact, the parameters they require are link probabilities, the type of causal relationships represented in the CPCS KB.

In the following discussion, we denote variables using upper-case letters (e.g., $X$) and instantiations of variables using lower-case letters (e.g., $x$). We consider all variables to be binary, where $\bar{x}$ means that $X$ is absent, and $x$ means $X$ is present.

The noisy OR is model of probabilistic causal influence between a binary effect variable and a set of binary variables that represent its causes. This repre-sentation was originally proposed by Pearl [1986] and independently by Peng and Reggia [1986].

For the noisy-OR network in Figure 2, let $V$ be the set of predecessor variables $\{D_1, ...D_m\}$. Let $V_X \subseteq V$ be the subset of predecessors of $X$ that are present and $\overline{V_X} \subseteq V$ be the subset of predecessors of $X$ that are absent. We assume that all predecessors are instantiated (thus, $V_X \cup \overline{V_X} = V$). We define $\underline{d_i}$ as the instantiation of $V$ in which $D_i$ is present and all other $D_j (j \neq i)$ are absent, or $\underline{d_i} \equiv \{d_i \wedge \cup_{j \neq i} \overline{d_j}\}$. We use $P(\mathbf{d})$ as shorthand for $P(\overline{d_1}, d_2, ..., d_m)$.

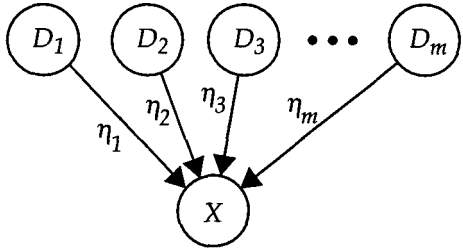

Figure 2: A noisy-OR network.

Let $\eta_i$, on the arc from $D_i$ to $X$, represent the *activation probability*, the probability that $X$ is present given that $D_i$ is present and all other predecessors are absent, i.e., $\eta_i = P(x|\underline{d_i})$.

Since the $D_i \in V_X$ are assumed to be causally independent, $X$ is absent only when all $D_i$ fail to cause $X$ to be present:

$$P(\overline{x}|V) = \prod_{i:D_i \in V_X} P(\overline{x}|\underline{d_i}),$$
$$= \prod_{i:D_i \in V_X} (1 - \eta_i). \tag{1}$$

If $X$ is binary, then we obtain from Eq. 1

$$P(x|V) = 1 - \prod_{i:D_i \in V_X} (1 - \eta_i). \tag{2}$$

Like any model, a BN is an incomplete representation of reality. We can use *leak events* to represent the missing variables that influence a finding. Each finding, or other variable with predecessors, has a corresponding leak event that represents all the possible events that could cause that finding to be present, other than those predecessor variables of the finding that are represented explicitly in the model. The leak probability for $X$, $\rho_X$, is equal to the probability that $X$ is present when all its explicitly modeled predecessors $\Pi(X)$ are absent:

$$\rho_X = P(l_X) = P(x|\overline{d_i} \quad \forall D_i \in \Pi(X)). \tag{3}$$

Thus, we can model the leak event like any other explicit cause $D_i$ of $X$. By incorporating a leak event into Eq. 3, we arrive at a formula for the leaky-noisy OR:

$$P(x|V) = 1 - (1 - \rho_X) \prod_{i:D_i \in V_X} (1 - \eta_i). \tag{4}$$

## 4    LEVEL REDUCTION OPERATION

This section outlines how we abstract the three-level CPCS-BN into a two-level network using an approximate and computationally efficient mapping. The mapping could be done exactly using arc reversal and barren node removal operations [Shachter, 1988], but that process is exponential in the size of the network. Our mapping will avoid these costly operations and provides bounds on the degree of entailed information loss that can be used to identify if more accurate mappings are necessary.

The level reduction process we describe can be applied to leaky noisy OR networks in which the nodes can be classified into a hierarchy of node types, such that the intermediate node classes can be abstracted away.

Reducing a network from three to two levels results in probabilities as shown in the right-hand-side network of Figure 3. The basic operation is to assign all paths between disease and finding nodes in the three-level network as edges in the resulting two-level network. Each edge in the two-level network takes on an activation probability given by the product of the activation probabilities on the path from which the edge was created.

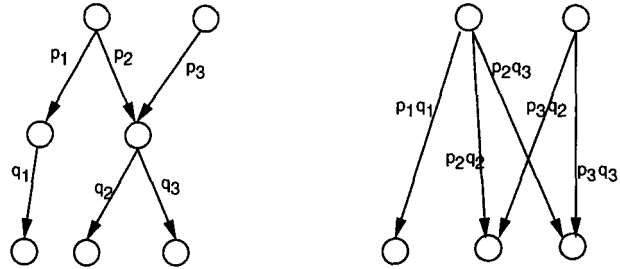

Figure 3: Network reduction using an approximate method

The following subsections outline our reduction mappings for activation probabilities and leak probabilities, followed by the algorithm for the reduction process.

### 4.1    LINK ACTIVATION PROBABILITY MAPPING

We define the initial three-level network to be comprised of a set $N$ of nodes. Levels of the network are indicated as node classes called $\mathcal{A}-$, $\mathcal{B}-$ and $\mathcal{C}-$nodes, starting from the top level. In other words, $\mathcal{A}$ represents disease nodes, $\mathcal{B}$ IPS nodes, and $\mathcal{C}$ finding nodes.

Consider a path connecting nodes $A \in \mathcal{A}$, $B \in \mathcal{B}$ and $C \in \mathcal{C}$. In particular, we define the $A$–$B$ edge to have activation probability denoted by $p_i$, and the $B$–$C$ edge to have probability denoted by $q_i$.[2] Using a simple approximation, we convert each two-edge path into a single-edge path. For a path with activation probabilities $p_i$ and $q_i$, the reduced activation probability, as shown in Figure 3 is defined as $p_i q_i$.

---

[2]There may be multiple IPS nodes in between a disease

This reduction is not exact, and we ignore all correlations. Thus, for example, the fork example in Figure 4 has a high level of correlation, which we are ignoring. In this figure, a correlation is induced when the three-level network edge with activation probability $q_1$ is "converted into" three edges in the two-level network, with activation probabilities $p_1q_1$, $p_2q_1$ and $p_3q_1$. These three resulting edges are correlated, since they "share" the activation probability $q_1$.

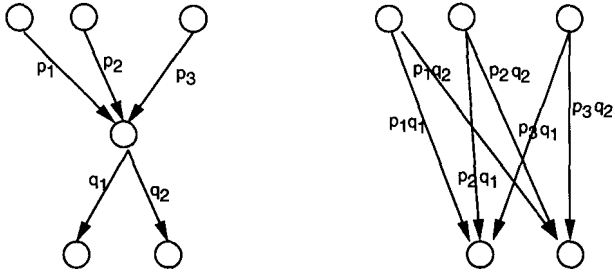

Figure 4: Subnetwork with a fork. We make an approximation by ignoring the correlation resulting from creating multiple links in the two-level network using a single edge in the three-level network. For example, the three-level network edge with activation probability $q_1$ is "converted into" three edges in the two-level network, with activation probabilities $p_1q_1$, $p_2q_1$ and $p_3q_1$.

We can reduce an $n$-level network, $n > 3$ to a 2-level network by recursively applying the operations described below.

For *diamond* subnetworks, such as that shown in Figure 5, we perform a mapping that "combines" the multiple paths between nodes $A$ and $C$ into a single edge. For example, the middle diagram in Figure 5 shows two edges between nodes $A$ and $C$ that are combined into a single edge. For a set of such edges with activation probabilities $p_1q_1, ..., p_kq_k$, the activation probability for the single "new" edge connecting nodes $A$ and $C$ is given by

$$\eta = 1 - \prod_j [1 - p_j q_j]. \tag{5}$$

## 4.2 LEAK ACTIVATION PROBABILITY MAPPING

Consider the leftmost portion of network shown in Figure 6. To reduce this to a two-level network, we need to create an updated leak variable for $C$, eliminating node $B$ and its leak node $L_B$ in the process. We use the leak updating process described in [del Favero *et al.*, 1995].

and finding node, so a path may have multiple edges.

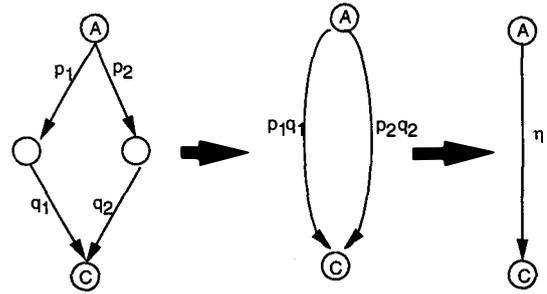

Figure 5: Reduction of diamond subnetwork. Multiple paths created in the reduced network are merged into a single path.

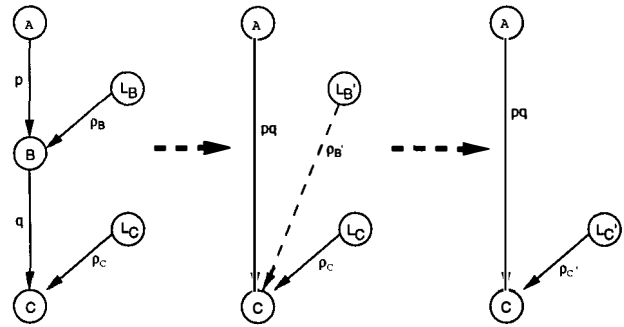

Figure 6: Leak node reduction.

The new leak activation probability is computed in two steps:

1. Update the leak probability for the path from $L_B$ to $B$ to $C$, as shown in Figure 6. The leak can activate $B$ when $A$ is absent, with a probability given by the leak probability $\rho_B$. We can reduce the path from $L_B$ to $B$ to $C$ to a single edge connecting the leak node for $B$ and node $C$. As shown in the center network of Figure 6, we will now have a new leak node $L'_B$ with leak probability given by $\rho_{B'} = \rho_B q$.

2. Use leak updating to absorb $L'_B$ and $L_C$ into a single new leak node, $L'_C$:

$$\rho_{C'} = 1 - (1 - \rho_{B'})(1 - \rho_C)$$
$$= 1 - (1 - \rho_B q)(1 - \rho_C).$$

## 4.3 LEVEL REDUCTION ALGORITHM

The level reduction algorithm applies the subnetwork reductions successively until a full reduction is complete. We define the reduction from 3 to 2 levels as follows:

Input: a 3-level network with node subsets $\mathcal{A}, \mathcal{B}, \mathcal{C}$ and edges $\mathcal{E}$.

Output: a 2-level network with node subsets $\mathcal{A}, \mathcal{C}$ and edges $\mathcal{E}'$.

*Level-reduction* $(\mathcal{A},\mathcal{B},\mathcal{C})$

1. For each disease $A_i \in \mathcal{A}$, IPS $B_j \in \mathcal{B}$, and finding $C_k \in \mathcal{C}$, if there is a link from $A_i$ to $B_j$ with activation probability $p_m$, and there is a link from $B_j$ to $C_k$ with activation probability $q_m$, create a new link $l_m$ from $A_i$ to $C_j$ with activation probability $p_m = p_i q_k$, and leak probability $\rho_m = 1 - (1 - \rho_j q_k)(1 - \rho_k)$.

2. For all sets of arcs $l_m$ that link $A_i$ to $C_k$, replace them all by a single link $l_n$ from $A_i$ to $C_k$ with link probability $p_n = 1 - \prod_m (1 - p_m)$ and with leak probability $\rho_n = 1 - \prod_m (1 - \rho_m)$.

Note that in CPCS-BN there are arcs joining pairs of IPS nodes such as joining $B_i$ to $B_j$, and these are first reduced by calling the algorithm on the path from disease $A$ to IPS $B_i$ to IPS $B_j$, obtaining an arc from $A$ to $B_j$.

## 4.4    EFFECTS OF LEVEL REDUCTION

We now describe one of the factors critical to understanding the effects of the level reduction operation: the *fan-in* and *fan-out* of an intermediate node. The fan-in of a node is the number of arcs terminating at the node and the fan-out is the number of arcs leaving the node. For example, the intermediate node in the subnetwork of Figure 4 has a fan-in of 3 and a fan-out of 2.

Consider a three-level network for our analysis. The first point to note is that for a node with fan-in of $m$ and fan-out of $n$, $m+n$ activation probabilities need to be specified, as opposed to $m \times n$ activation probabilities in the two-level representation of this subnetwork. Hence fewer probabilities need to be specified in the three-level version.

The second point is that the two-level version of the subnetwork will approximate the posterior probabilities assigned to diseases proportional to the values of fan-in and fan-out parameters. A fan-in and fan-out of 1 gives an exact reduction; when either parameter is greater than 1, the reduction becomes less accurate. Consider a subnetwork with fan-in of $m$ and fan-out of 1, where the IPS $I$ can take on value $i$, finding $F$ can take on value $f$, and diseases can take on values $\mathbf{d} = d_1, d_2, .., d_m$. As before, a disease-IPS activation probability is given by $p_i$, an IPS-finding activation probability by $q$, and a leak probability for node $X$ by $\rho_X$. The three-level network will have a probability of diseases given positive finding denoted by

$$P(\mathbf{d}|f) = \frac{P(f|\mathbf{d})P(\mathbf{d})}{P(f)}$$

$$= \frac{P(f|i)P(i|\mathbf{d})P(\mathbf{d})}{P(f)} + \frac{P(f|\bar{i})P(\bar{i}|\mathbf{d})P(\mathbf{d})}{P(f)}$$

$$= \{q(1 - \rho_I)[1 - \prod_{i=1}^{m}(1 - p_i)]$$

$$+ \rho_F \prod_{i=1}^{m}(1 - p_i)\} \frac{P(\mathbf{d})}{P(f)}.$$

The two-level network will have a probability of diseases given finding given by

$$P(\mathbf{d}|f) = \frac{P(f|\mathbf{d})P(\mathbf{d})}{P(f)}$$

$$= [1 - \prod_{i=1}^{m}(1 - p_i q)](1 - \rho_F)\frac{P(\mathbf{d})}{P(f)}.$$

The ratio $\mathcal{R}_1$ of three-level probability to two-level probability is equal to

$$\frac{\{q(1 - \rho_I)[1 - \prod_{i=1}^{m}(1 - p_i)] + \rho_F \prod_{i=1}^{m}(1 - p_i)\}\frac{P(\mathbf{d})}{P(f)}}{[1 - \prod_{i=1}^{m}(1 - p_i q)](1 - \rho_F)\frac{P(\mathbf{d})}{P(f)}}. \tag{6}$$

We can simplify this expression using the fact that $1 - \prod_{i=1}^{m}(1 - p_i) > \prod_{i=1}^{m}(1 - p_i)$ to obtain an approximate value for $\mathcal{R}_1$ of

$$\frac{(1 - \rho_I)\sum_i p_i - \prod_{i \neq j} p_i p_j + ... + p_1 p_2 \cdots p_m}{(1 - \rho_F)\sum_i p_i - q \prod_{i \neq j} p_i p_j + ... + q^{m-1}(p_1 p_2 \cdots p_m)}. \tag{7}$$

Assuming a value of $q < 1$, the two-level probability specified in the denominator of equation 7 is larger than that of the three-level probability when there is more than one disease. The relative difference between the numerator and denominator increases as $q$ decreases. In addition, the larger the values of the $p_i$, the less effect the value of $q$ has on the relative difference between numerator and denominator. This can be seen more clearly by looking at the case of two diseases, where we have:

$$\mathcal{R}_1 \simeq \frac{(1 - \rho_I)(p_1 + p_2 - p_1 p_2)}{(1 - \rho_F)(p_1 + p_2 - q p_1 p_2)}. \tag{8}$$

As an example, equation 7 predicts that for subnetworks of this type the two-level network will assign over-optimistically high posterior probabilities to multiple diseases (that are present) given positive evidence on the findings and activation probabilities of approximately the same size.

In an analogous fashion, we can derive a corresponding ratio for a network with fan-in of 1 and fan-out of $n$:

$$\mathcal{R}_2 = \frac{\frac{P(d)P(i|d)P(\mathbf{f}|i)}{P(\mathbf{f})} + \frac{P(d)P(\bar{i}|d)P(\mathbf{f}|\bar{i})}{P(\mathbf{f})}}{\frac{P(d)P(\mathbf{f}|d)}{P(\mathbf{f})}}$$

$$= \frac{\frac{P(d)p \prod_i P(f_i|i) + P(d) \prod_i \rho_{F_i}(1 - p)}{P(\mathbf{f})}}{\frac{P(d) \prod_i P(f_i|d)}{P(\mathbf{f})}}$$

$$= \frac{p \prod_i q_i + (1 - p) \prod_i \rho_{F_i}}{\prod_i p^i q_i}$$

$$\simeq \frac{1}{p^{n-1}}, \tag{9}$$

assuming that leak probabilities are small relative to activation probabilities. In this case, the two-level network underestimates the posterior probability assigned to $D$ given positive evidence for findings, and the degree of underestimation increases as $p$ gets smaller.

A network with fan-in and fan-out greater than 1 will thus be a combination of these effects. Note that the situation with both positive and negative evidence is significantly more complicated to analyze. We are currently examining these more general situations.

## 5 EXPERIMENTAL DESIGN

This section summarizes how we conducted our experiments. Key aspects of our design are the subnetworks and test cases used, and the measures of diagnostic performance employed. Our approach can be summarized as follows:

1. Select canonical full network ($n$-disease, binary)
2. Generate reduced (two-level) network
3. Generate test cases from full network
4. Perform inference on canonical network and reduced network to compute the probability assigned to the true diagnosis
5. Analyse results, using paired t-tests over the probability assigned to the true diagnosis for all the test cases.

### 5.1 SUBNETWORKS USED IN EXPERIMENTS

Because CPCS-BN is large and multiply-connected, it is impractical to perform inference using the entire network. If we wish only to compute the posterior probabilities of a small set of diseases, we would like to perform inference using only the subnetwork of the CPCS network that is relevant.

We can select subnetworks from the full CPCS-BN using the BN graphical tool Netview [Pradhan *et al.*, 1994], and then can export them for inference to an inference system such as IDEAL [Srinivas and Breese, 1990]. We can select subsets of the larger CPCS-BN, and can maintain correct probabilities in the subnetwork by appropriately updating leak probabilities, as described in [del Favero *et al.*, 1995].

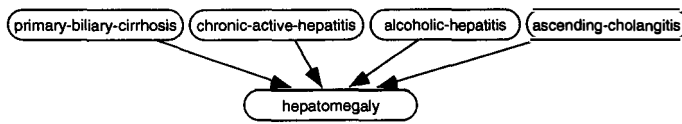

Figure 7: Predecessors of the node **hepatomegaly**.

When a subnetwork is saved, Netview updates the leak probabilities to take into account the missing diseases. In CPCS-BN, the node **hepatomegaly** has four disease predecessors shown in Figure 7. In subnetworks with fewer than four diseases, the updated leak probability for **hepatomegaly** is calculated based on this set of predecessors.

Table 1 summarizes the number of nodes of each type in the three subnetworks.

Table 1: Subnetworks used for experiments

| Network | Nodes | Diseases | IPSs | Findings |
|---------|-------|----------|------|----------|
| BN 2 | 42 | 2 | 2 | 38 |
| BN 3 | 146 | 3 | 46 | 97 |
| BN 4 | 245 | 4 | 80 | 161 |

### 5.2 TEST CASES

We needed far more test cases to estimate reliably the effects of the experimental manipulations on the diagnostic performance than the small number of cases available from real patient data. Accordingly, we generated sample test cases directly from the BNs themselves, generating findings according to the probabilities specified by the network, as described in [del Favero *et al.*, 1995].

Since we wanted to investigate how the amount of evidence affects sensitivity to the experimental manipulations, we generated cases with varying numbers of findings. The test cases, as initially generated, include values for all findings. To create harder cases with fewer findings, and also for greater medical realism, we created five cases from each initial case, by revealing the findings in five Phases, approximating the order in which findings would be revealed in a real medical consultation. Phases 1 and 2 are medical history and simple examination findings respectively, and phases 3 through 5 are tests of increasing expense and invasiveness.

### 5.3 MEASURES OF DIAGNOSTIC PERFORMANCE

We quantify diagnostic performance as the probability assigned by each network to each true diagnosis, averaged over the set of test cases. We analyze separately the probabilities assigned to each disease when present – that is, the true positive rate. Initially, we aggregate the results by phase so that we can see how performance varies by phase.

For statistical analysis of the results we used the paired t test to compare pairs of results, since we ran the same set of test cases under each condition. The posterior probabilities (true positive and false positive rates) often have very skewed distributions since they are constrained to be less than one. Accordingly, for statistical testing, we transformed them to the log-odds metric, which provides a distributions that appear more nearly normal.

## 6    RESULTS

We ran experiments on the three networks described in the previous section, and compared the average expected posterior probability across all diseases. Table 2 shows a sample of these results. This table shows that for every phase value, the 2-level network assigned a probability value to the true diagnosis almost identical to that assigned by the three-level network. These differences were statistically insignificant at the 97.5% confidence level.

A second point to notice is that the probability values increase with phase number, due to there being more evidence for the higher phase values.

Table 2: Summary results for networks: comparison of expected value of the true diagnosis versus phase for two-level and three-level network

**[2-disease network]**

| Phase | Two-level network | Three-level network |
|-------|-------------------|---------------------|
| 1     | 0.74983812        | 0.74983812          |
| 2     | 0.89429020        | 0.89434112          |
| 3     | 0.89415774        | 0.89500081          |
| 5     | 0.89521948        | 0.89562857          |

**[3-disease network]**

| Phase | Two-level network | Three-level network |
|-------|-------------------|---------------------|
| 1     | 0.177761          | 0.171843            |
| 2     | 0.491473          | 0.492253            |
| 3     | 0.579145          | 0.576442            |
| 4     | 0.651839          | 0.651839            |
| 5     | 0.672745          | 0.668560            |

**[4-disease network]**

| Phase | Two-level network | Three-level network |
|-------|-------------------|---------------------|
| 1     | 0.801361          | 0.799407            |
| 2     | 0.839976          | 0.838774            |
| 3     | 0.879373          | 0.876449            |
| 4     | 0.979329          | 0.979329            |
| 5     | 0.999710          | 0.999710            |

Table 3 compares the posterior probabilities for the three-disease network for two of the three diseases. Note that for ascending chalangitis, there are statistically insignificant differences between the probabilities for the two networks (at the 95% confidence level). In contrast, there is a statistically-significant difference (at the 95% confidence level) for the disease hepatitis-acute-viral. This is because the node for hepatitis-acute-viral has more IPS nodes in the path between it and its successor finding nodes than does the node for ascending chalangitis.

## 7    RELATED WORK

Several other approaches have been proposed to analyze the sensitivity of representation richness to KB

Table 3: Summary results for 3-disease network: comparison of expected value of the true diagnosis versus phase for two-level and three-level network, broken down by diseases

**[Ascending chalangitis]**

| Phase | Two-level network | Three-level network |
|-------|-------------------|---------------------|
| 1     | 0.24241785        | 0.24167455          |
| 2     | 0.27122661        | 0.27084166          |
| 3     | 0.39110151        | 0.39089065          |
| 4     | 0.47668803        | 0.47691644          |
| 5     | 0.97545444        | 0.97545614          |

**[hepatitis-acute-viral]**

| Phase | Two-level network | Three-level network |
|-------|-------------------|---------------------|
| 1     | 0.00312766        | 0.00351482          |
| 2     | 0.00502836        | 0.00335483          |
| 3     | 0.00492885        | 0.00472140          |
| 4     | 0.00725580        | 0.00462429          |
| 5     | 0.01177475        | 0.00449421          |

construction and inference costs, especially focusing on conditional independence assumptions. Fryback [1978], in a Bayesian model for medical diagnosis, showed empirically that large models with many inappropriate independence assumptions can have lower diagnostic accuracy than smaller models that do not include such inappropriate independence assumptions.

Provan [1994; 1993] studied the sensitivity of diagnostic accuracy to the richness and size of temporal BN models, using domain-dependent network reformulation methods. Network size provided a measure of the knowledge engineering and inference costs associated with the network. Provan found that diagnostic accuracy improved as a function of network complexity; however, if the model was penalized for its size, then the definition of best model was sensitive to both diagnostic accuracy and model size.

Wellman and Liu [1994] compared the results of various domain size abstractions for Bayesian networks for commuter-traffic routing. They used an anytime algorithm, and were primarily interested in the quality of the results given bounded time for inference. In this set of experiments we ignore the inference time, and focus solely on diagnostic performance.

## 8    CONCLUSIONS

In this paper, we have investigated how having intermediate nodes affects the diagnostic performance of a Bayesian belief network. In a series of experiments on a set of real belief subnetworks extracted from a large BN for medical diagnosis in liver and bile disease, we compared the effects on diagnostic performance of a two-level network consisting just of disease and finding nodes with that of the full network which models intermediate pathophysiological disease states as well.

We found that the two-level network assigned posterior probabilities to the the diseases that were virtually identical to those assigned by the three-level network: any differences are probably not of practical importance. Paired-t tests verified that these differences were, for the most part, statistically insignificant at the 95% level.

These results indicate that the intermediate nodes, for the network studied, do not make a statistically significant difference to the posterior probabilities assigned to diseases. Without such nodes, the network slightly overestimates the true-positive and false-positive probabilities for the larger networks. By studying the network topology of intermediate nodes, we have shown one factor that can be used to predict this degree of overestimation: the larger the number of parents and children that intermediate nodes have, the greater the overestimation. Since the larger networks have more densely-connected intermediate nodes, they will display larger differences of posteriors than smaller networks. Note that, given that the results for original and abstracted networks are virtually identical, it is not worthwhile studying more accurate network abstraction methods for this domain.

Our empirical results to date show that the additional effort required to build the CPCS KB from the two-level QMR KB does not lead to increased diagnostic accuracy. This is in accord with other experiments [del Favero et al., 1995] on CPCS-BN in which we showed that abstracting the CPCS-BN's quaternary-valued variables to binary-valued variables (as are present in QMR-DT) does not degrade diagnostic accuracy significantly.

One cannot apply these results to all domains, however. For example, in other domains constructing networks with intermediate nodes may be simpler and provide more accurate diagnoses than two-level networks. We plan to study similar abstractions in other domains, examine the role of intermediate nodes in inference speed, as well as analyze the theoretical properties of network abstractions in more detail.

**Acknowledgements**: Malcolm Pradhan provided assistance with running the experiments, and Max Henrion provided valuable commentary.

# Accounting for Context in Plan Recognition, with Application to Traffic Monitoring

**David V. Pynadath and Michael P. Wellman**
Artificial Intelligence Laboratory
University of Michigan
1101 Beal Avenue
Ann Arbor, MI 48109-2110 USA
{pynadath, wellman}@engin.umich.edu

## Abstract

Typical approaches to plan recognition start from a representation of an agent's possible plans, and reason evidentially from observations of the agent's actions to assess the plausibility of the various candidates. A more expansive view of the task (consistent with some prior work) accounts for the context in which the plan was generated, the mental state and planning process of the agent, and consequences of the agent's actions in the world. We present a general Bayesian framework encompassing this view, and focus on how context can be exploited in plan recognition. We demonstrate the approach on a problem in traffic monitoring, where the objective is to induce the plan of the driver from observation of vehicle movements. Starting from a model of how the driver generates plans, we show how the highway context can appropriately influence the recognizer's interpretation of observed driver behavior.

## 1 INTRODUCTION

The problem of *plan recognition* is to induce the plan of action driving an agent's behavior, based on partial observation of its behavior up to the current time. Deriving the underlying plan can be useful for many purposes—predicting the agent's future behavior, interpreting its past behavior, or generating actions designed to influence the plan itself. Researchers in AI have studied plan recognition for several kinds of tasks, including discourse analysis (Grosz & Sidner, 1990), collaborative planning (Huber & Durfee, 1993), and adversarial planning (Azarewicz *et al.*, 1989). These works have employed a great variety of reasoning techniques, operating on similarly various plan representations and adopting varied assumptions about observability.

The common theme underlying these diverse motivations and approaches is that the object to be induced is a *plan*, and that this plan is the cause of observed behavior. If there is anything special about the task of plan recognition as opposed to recognition in general, it must be due to special properties of plans: how they are constituted, and how they cause the behavior we observe and wish to predict, interpret, and influence.

In this paper, we focus on one of these special properties—the context in which the plan is generated—and how it can be exploited in the recognition process. Whereas most previous approaches have emphasized the relationship between plans and their observable effects,[1] we argue that it is equally necessary to consider evidence that would bear on which plan would have been appropriate for the agent to generate. We demonstrate this point through an example application in traffic monitoring, where the interpretation of an individual vehicle's action depends on the surrounding highway context. Our techniques for reasoning about plan-generation context are based on Bayesian networks, as part of a general Bayesian framework for plan recognition. This contribution can be considered a variant extension of the model of Charniak and Goldman (1993), and of the approach advocated by Huber et al. (1994).

## 2 PLAN RECOGNITION

### 2.1 TOWARDS A GENERAL BAYESIAN FRAMEWORK

One of the aims of our work is to elucidate the fundamental elements of plan recognition, and to develop a general Bayesian framework for approaches to this task. Achieving generality is complicated by the diversity of representations for plans and techniques for plan generation; therefore, we present the framework at multiple levels of specificity. The most abstract specification is designed to accommodate most conceivable versions of plan recognition, and by introducing further distinctions we taxonomize the ap-proaches.

---

[1] Although, as we point out in the discussion below, several of these approaches can also accommodate the sort of context information we are concerned with.

The framework for plan recognition is distinguished from uncertain reasoning in general by two special features of plans. First, plans are *structured linguistic objects*. Plan languages considered in AI research range from simple sequences of action tokens to general-purpose programming languages. In either case, the recognizer can and should exploit the structure of plans in inducing them from partial observations of the actions comprising the plan. Another way to say this is that plans are descriptions of action *patterns,* and therefore any general pattern-recognition technique is automatically a plan recognition technique for the class of plans corresponding to the class of patterns associated with the given technique.

The second special feature of plans is that they are *rational constructions*. They are synthesized by a rational agent with some beliefs, preferences, and capabilities, that is, a *mental state*. Knowing the agent's mental state and its rationality properties strongly constrains the possible plans it will construct. (The degree of constraint depends on the power of the rationality theory we adopt.) The rational origin of plans is what distinguishes plan recognition from pattern recognition. If the observations available include evidence bearing on the beliefs, preferences, and capabilities of the agent, then the recognizer should combine this with evidence from the observed actions in reasoning about the entire plan.

Our framework is *Bayesian* in that we start from a causal theory of how the agent's mental state causes its plan and executing its plan causes activity, and reason from observed effects to underlying causes. Our recognizer has uncertain *a priori* knowledge about the agent's mental state, the world state, and the world's dynamics, which can be summarized (at least in principle) by a probability distribution. It then makes partial observations about the world, and uses this evidence to induce properties of the agent and its plan.

The remainder of this section describes our framework in more detail. We demonstrate the utility of the framework by showing how extensions to the underlying conception of plans and planning generate corresponding extensions to plan recognition. Examples from our explorations of plan recognition in a highway traffic domain illustrate our application of the framework to a concrete problem.

## 2.2  PLANNING MODEL

We begin with a model of the planning agent operating in the world. As it begins planning, the agent has a certain mental state, consisting of its preferences (e.g., goals), beliefs (e.g., about the state of its environment), and capabilities (e.g., available actions). We assume the actual planning process to be some rational procedure for generating the plan that will best satisfy the agent's preferences based on its beliefs, subject to its capabilities. This plan then determines (perhaps with some uncertainty) the actions taken by the agent in the world.

Most plan-recognition work concentrates only on this last step, the relationship between a plan and the actions taken in the world. Typical approaches start from a representation of the possible plans, and prune the set of possibilities based on the actions observed. For example, Kautz (1986) connects plans and actions through event hierarchies, which place the plan at the top of a taxonomy of subplans and actions. Vilain (1990) presents a context-free grammar representation of these event hierarchies as an alternative model. Lin and Goebel (1991) restrict the constraint language, permitting use of a faster, specialized message-passing recognition algorithm.

Given the reduced set of possible plans that could explain the observations, the plan recognizer must apply some preference criterion for choosing among them. For instance, Kautz's approach prefers explanations that involve fewer plans. The algorithm of Lin and Goebel prefers plan scenarios that are more general. However, given two explanations containing the same number of plans, at the same levels of generality, neither algorithm has a basis for a choice either way. To borrow an example from Charniak and Goldman, suppose we hear that Jack packed a bag and went to the airport. Depending on the exact event hierarchy, neither algorithm may be able to decide whether Jack is in the process of taking a trip or conducting a terrorist bombing.

The average reader would probably not consider the latter possibility, since people are much more likely to take a trip than bomb an airplane. Charniak and Goldman account for this behavior in their recognition procedure by including prior probabilities on plans. This allows them to distinguish among equally possible, but unequally plausible explanations for observed activity. The recognition model of Carberry (1990), based on the Dempster-Shafer theory of evidential reasoning instead of Bayesian techniques, takes a similar approach by using threshold plausibility and difference levels of belief to distinguish among competing hypotheses. Similar distinctions could be supported in linguistic approaches as well, perhaps based on probabilistic grammars (Wetherell, 1980).

## 2.3  MENTAL STATE

In a particular case, we typically have information available to us that would augment these prior probabilities. For instance, we may know that Jack belongs to a terrorist organization, which would make the bombing explanation of his actions more plausible. To account for this sort of knowledge, the plan-recognition framework should accommodate all possible information about the agent's plan selection process, beginning with its mental state. We can break down an agent's mental state into three distinct com-